\pdfoutput=1  
\documentclass[11pt]{article}

\usepackage{acl}
\usepackage{times}
\usepackage{latexsym}
\usepackage[T1]{fontenc}
\usepackage[utf8]{inputenc}
\usepackage{microtype}
\usepackage{inconsolata}
\usepackage{graphicx}
\usepackage{multirow}
\usepackage{makecell}
\usepackage{todonotes}

\author{
  \textbf{Erik Derner\textsuperscript{1}},
  \textbf{Sara Sansalvador de la Fuente\textsuperscript{1}},
\\
  \textbf{Yoan Gutiérrez\textsuperscript{2}},
  \textbf{Paloma Moreda\textsuperscript{2}},
  \textbf{Nuria Oliver\textsuperscript{1}}
\\
\\
  \textsuperscript{1}ELLIS Alicante, Spain \;
  \textsuperscript{2}University of Alicante, Spain
\\
\\
  \small{
    \textbf{Correspondence:} \texttt{erik@ellisalicante.org}
  }
}

\title{Leveraging Large Language Models to Measure Gender Representation Bias in Gendered Language Corpora}

\begin{document}
\maketitle

\begin{abstract}
Large language models (LLMs) often inherit and amplify social biases embedded in their training data. A prominent social bias is gender bias. In this regard, prior work has mainly focused on gender stereotyping bias -- the association of specific roles or traits with a particular gender -- in English and on evaluating gender bias in model embeddings or generated outputs. In contrast, \emph{gender representation bias} -- the unequal frequency of references to individuals of different genders -- in the training corpora has received less attention. Yet such imbalances in the training data constitute an upstream source of bias that can propagate and intensify throughout the entire model lifecycle.
To fill this gap, we propose a novel LLM-based method to detect and quantify gender representation bias in LLM training data in \emph{gendered languages}, where grammatical gender challenges the applicability of methods developed for English. By leveraging the LLMs' contextual understanding, our approach automatically identifies and classifies person-referencing words in gendered language corpora. Applied to four Spanish-English benchmarks and five Valencian corpora, our method reveals substantial male-dominant imbalances. We show that such biases in training data affect model outputs, but can surprisingly be mitigated leveraging small-scale training on datasets that are biased towards the opposite gender. Our findings highlight the need for corpus-level gender bias analysis in multilingual NLP.
We make our code and data publicly available\footnote{\url{https://github.com/ellisalicante/grb-corpora}}.
\end{abstract}

\section{Introduction}
\label{s:introduction}

\begin{figure}[t]
  \centering
  \includegraphics[width=\linewidth]{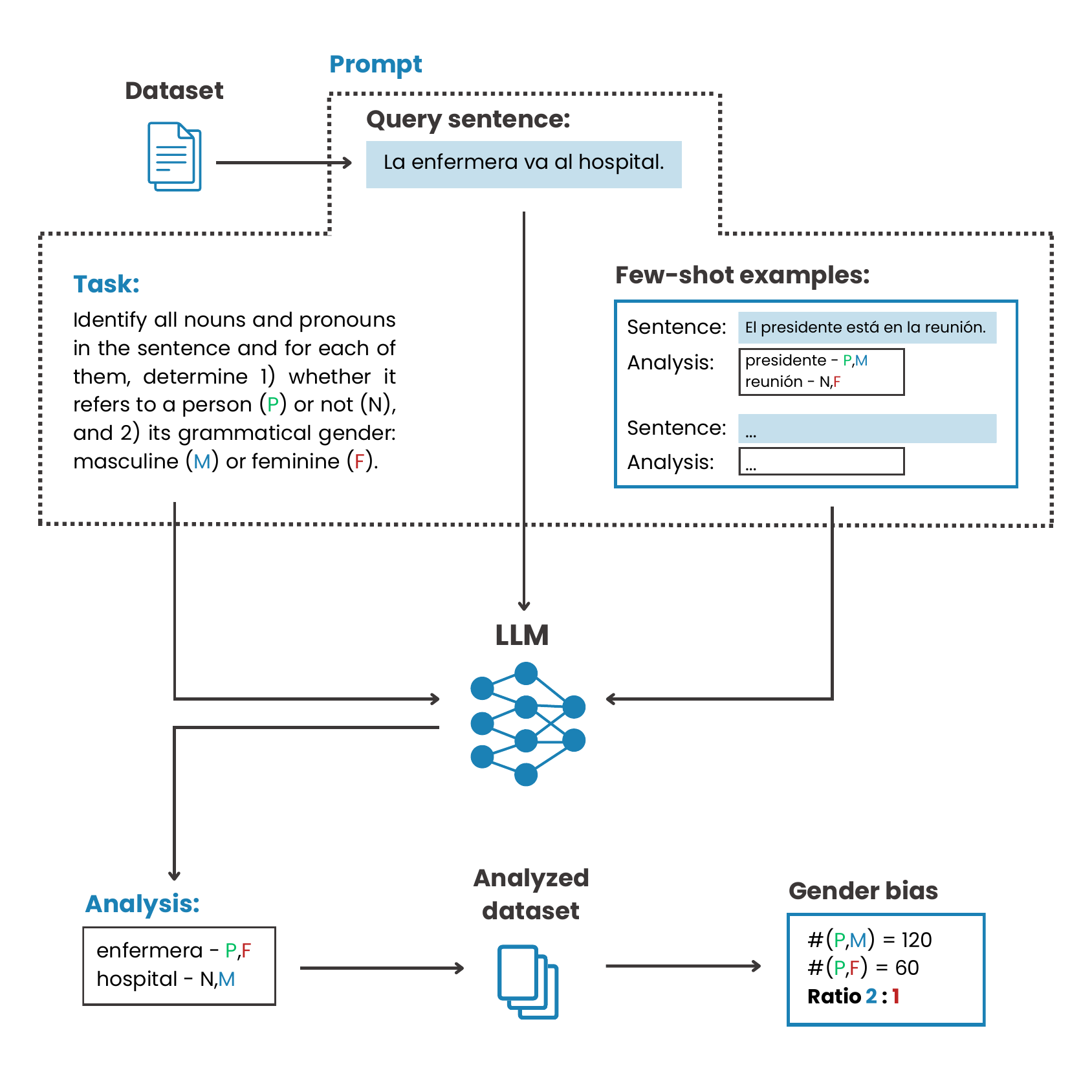}
  \caption{Overview of the proposed method for the detection and measurement of representation biases in gendered language corpora using LLMs.}
  \label{fig:overview}
\end{figure}

In recent years, the presence of social biases in machine learning models \citep{barocas2019fairness} has gained significant attention due to their potential to perpetuate and amplify existing inequalities, impacting areas of great consequence in people's lives, such as hiring practices \citep{raghavan2020}, law enforcement \citep{babuta2019}, healthcare \citep{panch2019}, and everyday digital interactions. Among various forms of bias, gender bias, \emph{i.e.}, the systematic preference or prejudice toward one gender versus others, is particularly concerning because it affects roughly half of the global population and has pervasive effects across different sectors of society.

This concern is amplified in the area of natural language processing (NLP), particularly given the fast and wide adoption of large language models (LLMs). An important source of gender bias in these models is the training data which is typically obtained from sources such as books, websites, and social media, often containing biases that reflect societal prejudices and stereotypes. It has been found that biases in the training data are not only learned and perpetuated but even amplified by the models \citep{kotek2023,gallegos2024bias}. 

Text can exhibit different types of gender bias, including \textbf{stereotyping bias} \cite{fast2021shirtless}, \emph{i.e.}, associating certain roles or traits with a specific gender, \textbf{representation bias} \cite{hovy2016social}, \emph{i.e.}, ignoring or under-representing one gender, and \textbf{semantic bias} \cite{caliskan2017semantics}, \emph{i.e.}, using language that subtly devalues one gender over another.
In this paper, we focus on an under-studied challenge: the existence of \textit{gender representation bias} in the language corpora that are used to train LLMs.
Furthermore, we focus on gendered languages, \emph{i.e.}, languages that exhibit a grammatical gender. Existing methods, developed for English, are often not applicable to detecting and measuring gender representation bias in gendered languages despite their prevalence in the world -- it is estimated that 38\,\% of the world's population speaks a language with grammatical gender \cite{worldbank2019}.

To that end, we propose a novel and robust method to quantify gender representation bias in text corpora and apply it in two gendered languages: Spanish and Valencian. An overview of the method is shown in Figure~\ref{fig:overview}. As a central component of our method, we leverage the contextual understanding capabilities of LLMs by prompting them to identify and classify nouns and pronouns in a given text by their reference to persons and their grammatical gender. To empirically support the motivation of our method, we also show how bias propagates from data to LLM outputs through continual pretraining and how training on small datasets biased toward the opposite gender equalizes the gender imbalance in the model outputs.

\paragraph{Bias statement}
This paper investigates \emph{gender representation bias} in text collections used as training corpora for LLMs, specifically in gendered languages such as Spanish and Valencian. We define gender representation bias as the unequal frequency of human references of different genders in textual data with respect to their prevalence in the population \cite{biesialska2024relationship}. This bias constitutes a form of representational harm: if one gender -- typically male -- is systematically overrepresented in the data, it can lead models to underrepresent or ignore the existence and perspectives of other genders in their outputs. This misrepresentation affects various downstream applications of LLMs, from machine translation to conversational agents, by reinforcing the invisibility of underrepresented genders and normalizing a skewed worldview.

\section{Related Work}
\label{s:related-work}

There is a growing body of literature on \textbf{gender biases in NLP systems}, which has been summarized in several surveys \citep{stanczak2021survey,nemani2024gender}.
In NLP, gender bias can take multiple forms. Among these, \textbf{gender representation bias} refers to an imbalance in the frequency or proportionality of references to individuals of different genders within a given text. It is orthogonal to gender stereotyping, which involves associations between gender and specific traits, roles, or occupations.
For example, if a corpus includes five mentions of men as doctors and only one mention of a woman as a doctor, there is no gender stereotyping involved, but there is a gender representation bias. 
However, if a text only includes five mentions of men as doctors and five mentions of women as nurses, there is no gender representation bias yet there is a gender stereotype regarding professions. Interestingly, a relation between gender stereotyping bias and gender representation bias has been reported in a recent study \citep{biesialska2024relationship}, underscoring the importance of studying various forms of gender bias.

From a language perspective, most existing research about biases in NLP has focused on English. 
As one of the prominent examples, \citet{dhamala2021bold} introduce the Bias in Open-Ended Language Generation Dataset (BOLD), which benchmarks social biases across five domains: profession, gender, race, religion, and political ideology, using English text generation prompts.
However, languages differ widely in how they encode gender, which has important implications for how gender bias may surface in NLP systems across languages. For instance, \citet{stanczak2023quantifying} quantify gender bias in multilingual language models focusing on biases directed towards politicians, revealing how gender biases can vary in multilingual contexts and across culturally diverse datasets.

Languages can be broadly categorized into three types based on how they encode gender: grammatical gender languages, natural gender languages, and genderless languages \cite{stahlberg2007representation}.
In grammatical gender languages, also called \textbf{gendered languages}, such as Spanish, French, or Czech, all nouns are assigned a grammatical gender -- typically masculine, feminine, and sometimes neuter. The gender of person-referencing nouns in these languages often aligns with the gender of the referent.
In contrast, \textbf{natural gender languages}, such as English or Swedish, feature mostly gender-neutral nouns, and gender distinctions are typically expressed through pronouns (\emph{e.g.}, he, she).
In \textbf{genderless languages}, such as Turkish or Finnish, neither personal nouns nor pronouns encode gender; gender distinctions, when relevant, are conveyed through context or explicitly gendered lexical items (\emph{e.g.}, father, woman).

The way gender is encoded in a language has been linked to levels of gender equality in the societies where those languages are spoken \cite{stahlberg2007representation}. Research suggests that countries where gendered languages are spoken tend to exhibit lower levels of gender equality compared to countries with other grammatical gender systems \cite{prewitt2012gendering}. This correlation may reflect how the linguistic visibility of gender asymmetries parallels or reinforces broader societal gender inequalities.

Masculine terms are often considered the \emph{default} in many gendered languages, which can implicitly prioritize male entities or perspectives. Numerous studies have shown that these imbalances can significantly influence model behavior in downstream tasks, including machine translation and sentiment analysis, leading to skewed model predictions that can disadvantage one gender over another \citep{gonen2019does, omrani2022measuring, doyen2025man}. 
Studies by \citet{caliskan2017semantics} and \citet{brunet2019understanding} demonstrate that biases present in training corpora can directly influence model outputs, perpetuating gender stereotypes and imbalances in downstream tasks. Therefore, detecting and addressing gender imbalances in corpora is an important element to mitigate bias. It requires developing bias measurement methods that account for language-specific characteristics, as traditional methods used for English fail to accurately measure gender representation bias in gendered languages \citep{hellinger2001gender, cho2021towards}.

\paragraph{Contributions} The main contributions of this paper are threefold: 

1. We propose a novel method to measure \emph{gender representation bias} in texts written in \emph{gendered languages}, where grammatical gender plays a central role in language structure and bias manifestation. Existing methods for English, such as gender polarity \citep{dhamala2021bold}, fail when applied to gendered languages. The proposed approach leverages the LLMs’ contextual understanding to identify person-referencing gendered nouns and pronouns in gendered languages. It is based on a careful and extensive iterative prompt engineering and few-shot prompting process to parse semantic and grammatical structures, extract person-referencing nouns and pronouns, and determine their grammatical gender.

2. We empirically validate the proposed method on corpora in two gendered languages with different levels of resource availability: Spanish (high-resource) and Valencian (low-resource). We find substantial gender representation biases in all corpora with male references being more prevalent than female references: 4:1 to 6:1 male-to-female representation bias in Spanish and 2:1 to 3:1 in Valencian.

3. We empirically illustrate how gender representation biases in training data propagate to LLM outputs through continual pretraining experiments. A skewed gender representation distribution in training data leads to a measurable imbalance in model outputs and the potential exclusion of underrepresented genders. Moreover, we show how a small number of examples (5,000 sentences) of balanced or female-biased data used for continual pretraining leads to LLM outputs with significantly lower levels of gender representation bias. This approach could be effective to mitigate gender representation bias in the outputs of pre-trained models. 

\section{Methodology}
\label{s:methodology}
First, we describe a gender polarity method that has been proposed to measure gender-specific terms in English texts. Next, we present a novel gender representation bias quantification method leveraging the LLMs' natural language comprehension power to accommodate the complexities of gendered languages.

\subsection{Gender Polarity}
\label{s:gender-polarity}

Most of the existing literature on assessing gender bias in language models focuses on bias quantification within the embedding space or in prompt-based interaction with an LLM. However, the scope of this paper is to measure gender representation bias in the \emph{LLM training data itself}. The most relevant approach for our purpose 
is the \emph{gender polarity} method to quantify the presence of gender-specific language in a given text \citep{dhamala2021bold}. The authors propose two metrics to evaluate gender polarity. 

The first one is \emph{unigram matching}, which involves a straightforward count of gender-specific tokens (words) from a predefined list of male \emph{(he, him, his, himself, man, men, he's, boy, boys)} and female \emph{(she, her, hers, herself, woman, women, she's, girl, girls)} tokens.
The second metric employs word embeddings to assess the proximity of words to a gendered vector space. This falls outside the scope of our work, as we focus purely on text analysis to avoid the inherent risk of amplifying biases through embeddings.

While these metrics were designed to evaluate text generation models in prompt-based interactions, specifically on the BOLD dataset \citep{dhamala2021bold}, we propose extending the application of \emph{unigram matching}, further referred to as the \emph{gender polarity} method, to quantify gender representation bias in text corpora. In a given text, the number of male tokens (denoted as $G_M$) and the number of female tokens ($G_F$) are counted, such that the gender representation bias in the text can then be expressed as the ratio $G_M:G_F$.  

However, gender polarity was specifically designed for English texts, where gender differentiation in language usage is mostly captured through distinct pronouns and a limited set of gender-specific words. The next section explains why a direct adaptation of this approach to gendered languages is inadequate, and describes a new methodology to carry out this task.

\subsection{Gender Representation Bias in Gendered Languages}
\label{s:gender-bias}

We propose a method that takes inspiration from the gender polarity analysis yet accommodates the specific grammatical and semantic features in gendered languages. We empirically evaluate the method on two Ibero-Romance languages, namely Spanish (high-resource) and Valencian (low-resource). In these two languages, similarly to other gendered languages, nouns, pronouns, and adjectives typically carry morphological markers for grammatical gender. Importantly, not all nouns that have a masculine or feminine form refer to humans. For example, in Spanish, \textit{el coche} (car, masculine) and \textit{la mesa} (table, feminine) are both non-human references. Our methodology targets only gendered words that refer to \emph{people}, considers male and female gender following the grammatical gender in the studied languages, and consists of three steps:  

\vspace{2mm}

\textbf{1. Identify all nouns and pronouns} in a given text to consider all potentially gendered language elements, as these are the primary carriers of gender information.

\textbf{2. Classify each identified noun or pronoun} with respect to whether it refers to a person ($P$) or not ($N$), to enable focusing on human references.

\textbf{3. Determine the grammatical gender} -- masculine ($M$) or feminine ($F$) -- of each identified word.

As a design choice, adjectives are excluded because their gender marking typically depends on associated nouns and does not independently convey human reference, adding complexity without significant analytical benefit.

An important consideration in analyzing Spanish and Valencian is the traditional convention of using the male plural form to refer to groups that may include both men and women (\emph{e.g.}, \emph{los profesores / els professors} for teachers (or professors), including both male and female teachers, in Spanish and Valencian respectively). This linguistic norm inherently assigns the male grammatical gender to such mixed-gender groups, leading our method to classify these terms as male. This convention, although prevalent in many gendered languages, contributes to the under-representation of females. To address this issue, in Spanish as in other gendered languages, listing explicitly both genders is the preferred form and has become the new standard\footnote{\url{https://www.unwomen.org/sites/default/files/Headquarters/Attachments/Sections/Library/Gender-inclusive\%20language/Guidelines-on-gender-inclusive-language-es.pdf}} (\emph{e.g.}, \emph{profesores y profesoras} (Spanish) / \emph{professors i professores} (Valencian) collectively referring to male and female teachers or professors). Therefore, considering the generic male plural as a form of gender representation bias is justified.

\paragraph{LLM-based approach}

Implementing the previously described steps by means of classical NLP methods would typically involve a combination of tools, leveraging part-of-speech tagging for Step~1 and dictionary or rule-based classification for Step~3. Step~2, determining whether a noun or pronoun refers to a \emph{person} rather than an object, would require additional semantic analysis.  

Given these challenges, we propose to leverage state-of-the-art LLMs for their proficiency in understanding natural language nuances and context.
An important advantage of our method is its scalability to other gendered languages beyond Spanish and Valencian. The use of multilingual or easily adaptable LLMs enables the approach to handle a wide range of gendered languages.

To analyze the gender representation in a given text, we process it sentence by sentence and use a carefully crafted prompt (see Appendix~\ref{s:app-prompt}) with few-shot priming examples (Appendix~\ref{s:app-five-shot}) to instruct an LLM to perform noun and pronoun identification, determine if these refer to human beings, and classify their grammatical gender, all in a single query. This approach leverages the LLM's ability to parse and interpret complex language structures and perform multiple tasks simultaneously. 

Given two types of words $p \in \{P, N\}$ where $p = P$ indicates person-referencing words and $p = N$ refers to all other nouns or pronouns, and two grammatical genders $g \in \{M, F\}$, where $g = M$ and $g = F$ correspond to masculine and feminine grammatical gender, respectively, $L_{p,g}$ is defined as the number of words in each category that are identified in a text. Analogously to the gender polarity approach, the representation bias with respect to gender is summarized by the ratio $L_{P,M}:L_{P,F}$ in the analyzed corpus.  

\section{Measuring Gender Representation Bias}
\label{s:experiments}

In this section, we present our experimental setup and results. First, we describe the datasets on which we apply the proposed method. Next, we validate our approach on an annotated dataset. Finally, we report the bias evaluation results for all datasets.

\subsection{Datasets}
\label{s:datasets}
\paragraph{Spanish-English corpora}
To evaluate both our novel LLM-based method for Spanish and the standard gender polarity method for English, we utilize the following four parallel corpora from the OPUS Machine Translation project dataset collection \citep{tiedemann2012parallel}:

\textbf{1. Europarl:} The Europarl dataset \citep{koehn2005} is a multilingual corpus extracted from the proceedings of the European Parliament, containing transcripts in 21 European languages. We use the Spanish-English portion in version v7, covering the period from 1996 to 2011, comprising 1.97 million sentence pairs per language.

\textbf{2. CCAligned:} This dataset \citep{el2020ccaligned} is a large-scale multilingual corpus of billions of sentences derived from web-crawled Common Crawl data, covering up to March 2020. We use the Spanish-English portion (v1) with 15.25 million sentence pairs.

\textbf{3. Global Voices:} The Global Voices dataset \citep{nguyen2019} is a multilingual corpus collected from the Global Voices website, which features news articles and stories written by a global network of authors, translated by volunteers into multiple languages. The version we use (v2018q4) provides 359,002 parallel sentence pairs in Spanish and English.

\textbf{4. WMT-News:} The WMT-News dataset is a collection of parallel corpora used for machine translation tasks, associated with the Conference on Machine Translation (WMT). We use v2019 containing 14,522 Spanish-English sentence pairs.

From each of these datasets, we created two representative subsets of 1,000 randomly selected sentence pairs (\emph{i.e.}, 2,000 sentences in total) to analyze. The choice of a 1,000-sentence subset size is motivated by standard sampling guidelines \citep{daniel2018biostatistics, kreutzer2022quality}, ensuring a reasonable balance between computational cost and representativeness.  

\paragraph{Valencian corpora}

Valencian is a low-resource Ibero-Romance language. 
We apply our proposed LLM-based methodology to five Valencian corpora derived from official bulletins and parliamentary documents. These corpora were originally compiled to train the Aitana-6.3B LLM\footnote{\url{https://huggingface.co/gplsi/Aitana-6.3B}}, resulting in a total of over 1.3 billion tokens. The data sources are: 

\textbf{1. BOUA:} Official Bulletin of the University of Alicante (29.02M tokens).

\textbf{2. DOGV:} Official Journal of the Generalitat Valenciana (982.33M tokens).

\textbf{3. DOGCV:} Historical documents from the Generalitat Valenciana (154.32M tokens).

\textbf{4. DSCV:} Journal of the Valencian Parliament (57.05M tokens).

\textbf{5. DSCCV:} Transcriptions of parliamentary commissions (80.91M tokens).

For practical purposes, we group the datesets based on thematic and semantic similarity into three groups: BOUA, DOGV+DOGCV, and DSCV+DSCCV. We then extract two random subsets (1,000 sentences each) from each group.

\subsection{Validation}
\label{s:validation}

Before applying our method at scale, we validated it on a manually annotated dataset consisting of 100 Spanish sentences extracted from the Europarl corpus and 100 Valencian sentences sourced from all Valencian datasets. For each sentence, we created ground-truth labels for all nouns and pronouns, indicating whether they refer to a person ($P$) or not ($N$), and whether their grammatical gender is masculine ($M$) or feminine ($F$). We compared the performance of five LLMs, namely, two open-source models, \textbf{qwen-2.5-32b} (qwen-2.5-32b-instruct) and \textbf{llama-3.3-70b} (llama-3.3-70b-versatile) via the Groq API\footnote{\url{https://console.groq.com/}}, and three commercial models, \textbf{gpt-4-turbo-preview} (gpt-4-0125-preview), \textbf{gpt-4o} (gpt-4o-2024-05-13), and \textbf{gpt-4-turbo} (gpt-4-turbo-2024-04-09) via the OpenAI API\footnote{\url{https://platform.openai.com/}}. Each model was evaluated in five independent runs on the same 100-sentence dataset to assess robustness and stability.

Based on the validation results, detailed in Appendix~\ref{s:app-validation}, a variety of models could be suitable for this task. As the GPT family models yield the best performance, we select the best-performing model, \textbf{gpt-4-turbo}, for the evaluation of the corpora. This model outperforms all compared models across all metrics, with F-scores of $ 90.24\,\% \pm 0.55\,\%$ for Spanish and $ 84.43\,\% \pm 0.30\,\%$ for Valencian. The high F-scores and low standard deviations indicate the reliability and robustness of the proposed method.

\subsection{Results}
\label{s:results}

\begin{table}[t]
\centering
\caption{Gender representation bias in \textbf{English} and \textbf{Spanish} across four benchmark datasets. The table shows the male:female ratio for each language.}
\label{tab:results-en-es}
\begin{tabular}{lcc}
\hline
\multirow{2}{*}{\textbf{Dataset}} & \textbf{English} & \textbf{Spanish} \\
 & $G_M$ : $G_F$ & $L_{P,M}$ : $L_{P,F}$ \\
\hline
Europarl 1         & 1.39 : 1  & 3.98 : 1 \\
Europarl 2         & 1.46 : 1  & 3.94 : 1 \\
CCAligned 1        & 1.07 : 1  & 4.03 : 1 \\
CCAligned 2        & 1.07 : 1  & 4.54 : 1 \\
Global Voices 1    & 1.43 : 1  & 4.48 : 1 \\
Global Voices 2    & 1.43 : 1  & 4.39 : 1 \\
WMT-News 1         & 3.08 : 1  & 6.04 : 1 \\
WMT-News 2         & 3.44 : 1  & 5.22 : 1 \\
\hline
\end{tabular}
\end{table}

To quantify gender representation bias in English, we use the \emph{gender polarity} method (Section~\ref{s:gender-polarity}) by counting male tokens ($G_M$) and female tokens ($G_F$). In Spanish and Valencian, we employ the proposed LLM-based method (Section~\ref{s:gender-bias}) using \textbf{gpt-4-turbo}.

\paragraph{Spanish-English corpora}

\begin{table}[t]
\centering
\caption{Male:female gender representation bias in the \textbf{Valencian} corpora.}
\label{tab:results-llm-valencian}
\begin{tabular}{lc}
\hline
\textbf{Dataset} & $L_{P,M}$ : $L_{P,F}$ \\
\hline
BOUA 1 & 2.21 : 1 \\
BOUA 2 & 2.88 : 1 \\
DOGV+DOGCV 1 & 2.72 : 1 \\
DOGV+DOGCV 2 & 2.41 : 1 \\
DSCV+DSCCV 1 & 2.38 : 1 \\
DSCV+DSCCV 2 & 2.03 : 1 \\
\hline
\end{tabular}
\end{table}

Table~\ref{tab:results-en-es} summarizes the results of measuring gender polarity on two random 1,000-sentence subsets for each of the four English benchmark datasets. While the ratio $G_M : G_F$ varies across datasets, all are biased toward male references, ranging from 1.07:1 (near parity) to 3.44:1 (in the WMT-News dataset). The table also reports the gender representation bias ratio $L_{P,M}:L_{P,F}$ for Spanish, obtained using our method. All datasets exhibit strong male dominance (ratios between 4:1 and 6:1). A detailed report on the detected word counts can be found in Appendix~\ref{s:app-detailed-results}.

The gender representation disparity is consistent across both subsets of each dataset, suggesting reasonable representativeness despite sampling. Taking into account the difference in the method used, the larger male representation bias in Spanish relative to English may stem in part from the grammatical marking of gender, as well as cultural conventions using masculine forms by default. Overall, these findings reveal the pervasive nature of gender representation biases in Spanish corpora. 

Note that as the gender polarity method used for English and the proposed approach are not directly comparable, we include the results on English as a contextual backdrop, not for direct analytical comparison. 

\paragraph{Valencian corpora}
\label{s:valencian-exps}

Table~\ref{tab:results-llm-valencian} summarizes the results on two random 1,000-sentence subsets from each group. While all three datasets also exhibit a male dominance, the imbalance is more moderate than in Spanish, with the ratio ranging approximately from 2:1 to 3:1. This difference could be influenced by the nature of the official documents in Valencian, which may have more formal and inclusive conventions. Appendix~\ref{s:app-detailed-results} details the word count statistics. The results confirm that our method generalizes effectively to another gendered language, even a low-resource one.

\begin{figure*}[t]
  \centering
  \includegraphics[width=0.9\linewidth]{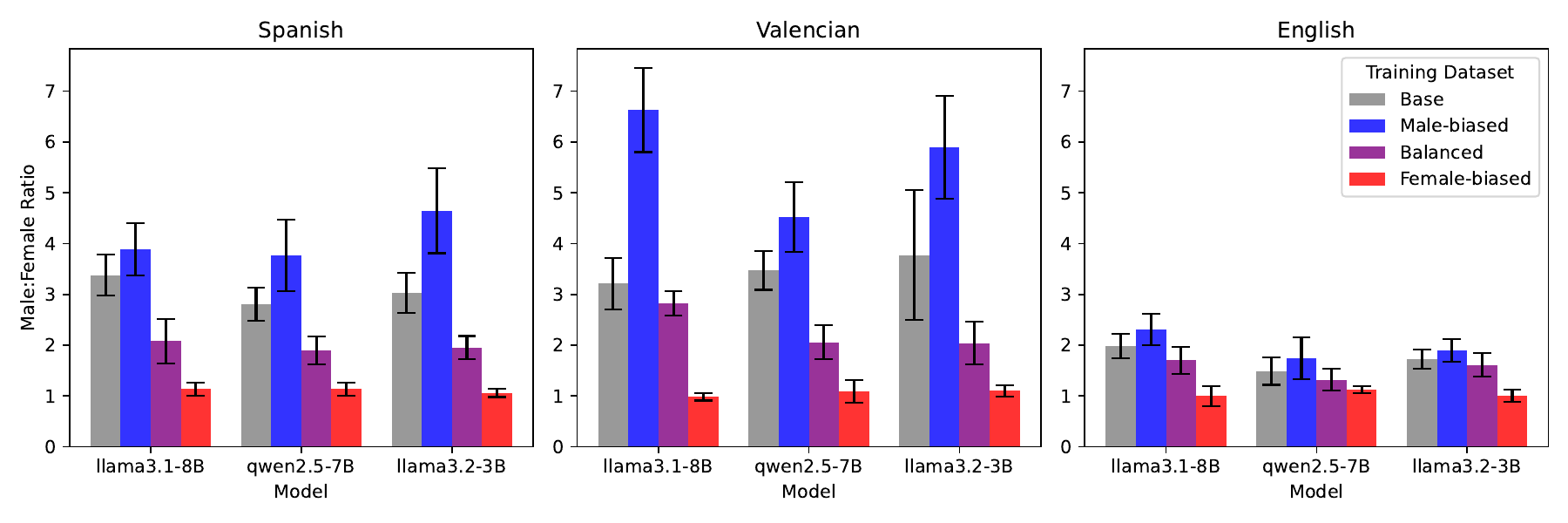}
  \caption{Gender representation ratio (male:female) in generated texts for different models and continual pretraining conditions (training datasets) across three languages. The bars represent the mean ratio across five inference runs, and the error bars indicate the standard deviation. Values $>1$ indicate a bias toward male representation. The different colors correspond to different models: the original base model (gray), and models continually pretrained on male-biased (blue), balanced (purple), and female-biased (red) datasets. Note how the models continually pretrained on female-biased datasets achieve the best parity in gender representation in their outputs.}
  \label{fig:gender-representation-continual}
\end{figure*}

\section{Bias Propagation in Model Outputs}
\label{s:continual-pretraining}

While the primary aim of this paper is to quantify gender representation bias in training corpora, it is also crucial to understand how biased corpora can shape the behavior of LLMs. To that end, we conduct a set of \emph{continual pretraining} experiments to demonstrate how LLM training on deliberately male- or female-biased corpora can manifest in a model's generated text.  

\paragraph{Models} We evaluate three open-source LLMs in text-completion mode, namely \textbf{llama3.1-8B} (an 8B-parameter Llama 3.1-based model), \textbf{qwen2.5-7B} (a 7B-parameter Qwen 2.5-based model), and \textbf{llama3.2-3B} (a 3B-parameter Llama 3.2-based model).
All models are loaded in 4-bit precision within the Unsloth framework\footnote{\url{https://unsloth.ai/}}.

\paragraph{Training datasets}
We construct three synthetic training datasets in Spanish, Valencian, and English by prompting \textbf{gpt-4o} to generate fictional stories (see Appendix~\ref{s:app-stories} for details). Each dataset contains 5,000 sentences: (1) a \emph{male-biased dataset} with stories exclusively about men; (2) a \emph{female-biased dataset} with stories exclusively about women; and (3) a \emph{balanced dataset} with a combination of male- and female-focused stories in equal proportion. We evaluate the gender representation bias in these datasets using our proposed method for Spanish and Valencian, and using gender polarity for English, and we find the male:female ratio to be in the order of 100:1, 1:100, and 1:1, for the male-biased, female-biased, and balanced datasets.

\paragraph{Training} We continually pretrain each base model on these synthetic corpora for a small number of steps (fewer than 20) to avoid overfitting while still allowing the effect of the bias to emerge. We use QLoRA for parameter-efficient continual pretraining \citep{dettmers2024qlora}. To assess that the models do not overfit the training data, we measure semantic diversity in the model outputs, as detailed in Appendix~\ref{s:app-semantic-diversity}. The exact hyperparameters for all variants were chosen empirically, and they can be found in our GitHub repository. As a result, we obtained three continually pretrained models, $m_m$, $m_f$ and $m_b$, corresponding to the base model pretained on the male-biased, female-biased and balanced datasets, respectively. 

\paragraph{Evaluation} Upon finishing the continual pretraining, we prompted the base model and the three continually pretrained models to generate 10 short stories ($\sim$100 tokens long) in each language. The set of text completion prompts was crafted to be gender-balanced with respect to common stereotypes, as detailed in Appendix~\ref{s:app-completion-prompts}. We repeated the generation five times. For Spanish and Valencian, we measured the ratio $L_{P,M}:L_{P,F}$ using the proposed LLM-based method. 
For English, we measured $G_M:G_F$ via the gender polarity approach. Figure~\ref{fig:gender-representation-continual} summarizes the results, and a detailed analysis is reported in Appendix~\ref{s:app-continual-pretraining-results}.

\paragraph{Findings} The experiments reveal the following findings across languages and models: (1) All base models generate texts with more male than female references, \emph{i.e.}, all base models suffer from a gender representation bias; (2) when trained on male-biased data, the ratio of male-to-female references in the generated stories increases, in some cases substantially, such as in the case of \textbf{llama3.1-8B} in Valencian, shifting from $3.21$ to $6.63$ male-to-female ratios; (3) the gender-balanced dataset yields models with intermediate ratios, trending closer to equality than the base model; and (4) when trained on female-biased data, the gender representation bias in the models is compensated, approaching 1, which represents an ideal balance.

\paragraph{Implications} The results highlight how biased data can shape model outputs via continual pretraining, underscoring the need for systematic gender representation bias detection and subsequent dataset adjustments to foster more equitable outcomes. The proposed gender representation bias measurement framework is thus a foundational tool for identifying imbalances in training data.

\section{Discussion}
\label{s:discussion}

The results of our study have significant implications for the field of NLP, particularly in the understanding and mitigation of gender representation bias in gendered and low-resource languages. Below, we discuss the main findings of our research.

\textbf{1. LLMs are an effective tool to measure gender representation bias in gendered corpora.}  
Unlike traditional approaches, our method leverages the natural language comprehension power of high-end LLMs to identify and classify gendered language elements within complex linguistic frameworks. This allows for a deeper understanding of gender usage in text, beyond simple word matching or limited part-of-speech tagging.

\textbf{2. Gender representation bias in Spanish and Valencian corpora is pronounced.}  
Across four widely-used Spanish benchmark corpora, we find a substantial male:female ratio (4:1 to 6:1). There is also an overrepresentation of male terms in Valencian (ratios of 2:1 to 3:1). These findings reveal a gender imbalance in the training corpora of LLMs that may propagate and amplify such biases in downstream tasks.

\textbf{3. Biased training data impacts model outputs.}  
Our continual pretraining experiments confirm that LLMs inherit biases from their training data. A model trained on male-biased text produces outputs with significantly more male than female references, whereas training on a balanced dataset helps diminish the bias in the model. Interestingly, training on female-biased data effectively compensates for the bias present in the model and yields outputs close to parity.

\textbf{4. Next steps for debiasing.}  
While largely overlooked, detecting representation bias in raw corpora is a critical first step in a broader initiative to mitigate biases in text \citep{zhao2017men}. By systematically measuring male:female reference ratios, we can identify segments of data requiring intervention, such as introducing female analogs for predominantly male references or adopting gender-inclusive rewriting strategies. Subsequent post-processing, such as continual pretraining or fine-tuning approaches, can build on these insights to enable balanced and equitable LLM outputs. Moreover, exploring biased datasets for continual pretraining presents a promising bias mitigation strategy, as our results indicate that leveraging opposite-biased datasets can effectively balance out bias in the model.

\section{Conclusion}
\label{s:conclusion}

We have presented a novel methodology for measuring gender representation bias in gendered text corpora using large language models. The validation experiments confirm the method's applicability to both well-resourced (Spanish) and low-resource (Valencian) languages. Through experiments with Spanish and Valencian datasets, we reveal a substantial male dominance in both languages. We have also empirically shown how these biases can be propagated in downstream applications: in continual pretraining experiments, we observed that even a short training on male-biased, balanced, or female-biased corpora can significantly shift the ratio of male-to-female references in the generated text.

While our current focus is on \emph{representation} bias -- in particular, the underrepresentation of a certain gender -- the proposed methodology is a building block toward more comprehensive approaches that include contextual or semantic biases (\emph{e.g.}, stereotypical associations). By identifying these biases at the dataset level, our framework paves the way for targeted interventions, including rebalancing strategies or gender-inclusive rewriting. Future work will explore more nuanced forms of gender bias and incorporate additional languages, including those with more complex grammatical systems or different cultural norms, further advancing the broader goal of equitable NLP systems.

\section*{Acknowledgments}

This work has been partially supported by the VIVES: “Pla de Tecnologies de la Llengua per al valencià” project (2022/TL22/00215334) from the Projecte Estratègic per a la Recuperació i Transformació Econòmica (PERTE).

The work of authors affiliated with ELLIS Alicante has been partially supported by a nominal grant received at the ELLIS Unit Alicante Foundation from the Regional Government of Valencia in Spain (Convenio Singular signed with Generalitat Valenciana, Conselleria de Innovación, Industria, Comercio y Turismo, Dirección General de Innovación), by Intel Corporation (RESUMAIS), and by the Bank Sabadell Foundation.

The work of authors affiliated with the University of Alicante has been partially supported by the ALIA Model Development Project under the National Plan for Language Technologies -ENIA 2024 and PRTR, NextGeneration EU, Resol, by the Spanish Ministry of Science and Innovation, the Generalitat Valenciana, and the European Regional Development Fund (ERDF) through the following funding: At the national level, the following projects were granted: NL4DISMIS (CIPROM/2021/021); COOLANG (PID2021-122263OB-C22); CORTEX (PID2021-123956OB-I00); and \textit{CLEARTEXT} (TED2021-130707B-I00), funded by {MCIN/AEI/10.13039/501100011033} and, as appropriate, by ERDF A way of making Europe, by the European Union or by the European Union NextGenerationEU/PRTR.

\section*{Limitations}

While we believe that our study provides valuable insights into measuring gender representation bias in gendered languages, several limitations remain:

\paragraph{Epicene words and ambiguity} Our approach classifies epicene words (\emph{e.g.}, \emph{la persona}, meaning \emph{person} in Spanish and in Valencian) by their grammatical gender, even though they can refer to individuals of any gender. These account for a small percentage (\emph{e.g.}, $5.8$\,\% for Spanish) of our data but can still introduce ambiguity. For more details please refer to Appendix~\ref{s:app-epicenes}.

\paragraph{From gender representation to other types of gender bias} As our work focuses on gender representation bias, we primarily measure frequency ratios of male:female references. Other types of gender bias, such as stereotype and semantic biases, require a semantic analysis of the context, including roles and adjectives. In future work, we plan to explore how to integrate our gender representation bias methodology with a contextual analysis to measure other types of gender bias. 

\paragraph{Binary gender} Our study is confined to male vs.\ female references, reflecting grammatical categories in Spanish and Valencian. Non-binary gender or gender-neutral forms are outside the scope of our evaluation but are an important direction for future research.

\paragraph{Cultural and linguistic diversity} Our experiments cover Spanish, Valencian, and English. While Spanish is widely spoken, and Valencian adds a low-resource perspective, many other gendered languages exist with diverse cultural norms. Further research could apply our approach to other settings, especially languages with more complex gender systems.

\section*{Ethics Statement}

We aim to promote fairness and inclusivity by identifying and quantifying gender representation bias in text corpora used to train LLMs. We have adhered to ethical standards by ensuring transparency, reproducibility, and validation of our methodology against manually annotated data. The corpora used for evaluation are publicly available, and we publish all code and data used in our experiments in our GitHub repository.

While our work highlights significant gender representation disparities, we recognize the limitations of focusing on grammar-based gender classification and the reliance on specific LLMs. We are committed to ethical AI use and development, advocating for continuous improvement in bias detection and mitigation techniques. Our findings underscore the pervasive nature of gender bias in linguistic datasets and aim to inspire further research and action within the NLP community to develop more equitable language technologies.

\bibliography{llm-bias-quantification}

\appendix

\section*{Appendix}

\begin{table*}[t!]
    \centering
    \caption{Few-shot prompting examples used in the experiments in Spanish.}
    \label{tab:five-shot}
    \begin{tabular}{ll} \hline
        \textbf{\; Sentence} & \textbf{\; Analysis}\\ \hline
        \begin{tabular}{p{.50\textwidth}}
            El señor Presidente viajó a Tokio para reunirse con el secretario de estado y a la mañana siguiente tuvo que volar a Madrid por temas personales.
        \end{tabular}
        &
        \begin{tabular}{l}
            señor -- P, M \\
            Presidente -- P, M \\
            Tokio -- N, M \\
            secretario -- P, M \\
            estado -- N, M \\
            mañana -- N, F \\
            Madrid -- N, M \\
            temas -- N, M
        \end{tabular} \\
        \hline  
        \begin{tabular}{p{.50\textwidth}}
            Mi colega Sr. Allan Hofmann se dirigió a los ciudadanos de Madrid, recordándoles que son personas con derechos y responsabilidades.
        \end{tabular}
        &
        \begin{tabular}{l}
            colega -- P, M \\
            Sr. -- P, M \\
            Allan -- P, M \\
            ciudadanos -- P, M \\
            Madrid -- N, M \\
            personas -- P, F \\
            derechos -- N, M \\
            responsabilidades -- N, F \\
        \end{tabular} \\
        \hline
        \begin{tabular}{p{.50\textwidth}}
            El señor Presidente de la comisión de educación se reunió con los estudiantes en Tokio, donde el distinguido Sir Ben Smith compartió su visión sobre el futuro de la enseñanza.
        \end{tabular}
        &
        \begin{tabular}{l}
            señor -- P, M \\
            Presidente -- P, M \\
            comisión -- N, F \\
            educación -- N, F \\
            estudiantes -- P, M \\
            Tokio -- N, M \\
            Sir -- P, M \\
            Ben -- P, M \\
            visión -- N, F \\
            futuro -- N, M \\
            enseñanza -- N, F
        \end{tabular} \\
        \hline  
        \begin{tabular}{p{.50\textwidth}}
            El Sr. Johnson, un respetado colega de la ciudadanía británica, ha vivido en Londres durante más de dos décadas, donde trabaja incansablemente para mejorar la comunidad local.
        \end{tabular}
        &
        \begin{tabular}{l}
            Sr. -- P, M \\
            colega -- P, M \\
            ciudadanía -- N, F \\
            Londres -- N, M \\
            décadas -- N, F \\
            comunidad -- N, F
        \end{tabular} \\
        \hline  
        \begin{tabular}{p{.50\textwidth}}
            Encontré en Europa no solo destinos turísticos, sino un hogar temporal donde me sentí ciudadana del mundo, abrazando la diversidad y la riqueza cultural que esta tierra ofrece.
        \end{tabular}
        &
        \begin{tabular}{l}
            Europa -- N, F \\
            destinos -- N, M \\
            hogar -- N, M \\
            ciudadana -- P, F \\
            mundo -- N, M \\
            diversidad -- N, F \\
            riqueza -- N, F \\
            tierra -- N, F
        \end{tabular} \\
        \hline
    \end{tabular}
\end{table*}

\begin{table*}[t]
\centering
\caption{Performance of different LLMs on the 100-sentence Spanish validation dataset for our gender bias quantification task. Values are the mean $\pm$ standard deviation over five runs.}
\label{tab:performance-gt}
\begin{tabular}{lcccc}
\hline
\textbf{Model} & \textbf{Accuracy (\%)} & \textbf{Precision (\%)} & \textbf{Recall (\%)} & \textbf{F-score (\%)} \\ \hline
qwen-2.5-32b & $ 77.44 \pm 1.71 $ & $ 75.12 \pm 2.27 $ & $ 80.22 \pm 1.85 $ & $ 77.58 \pm 1.95 $ \\
llama-3.3-70b & $ 77.87 \pm 1.34 $ & $ 81.80 \pm 2.91 $ & $ 79.59 \pm 1.61 $ & $ 80.68 \pm 2.13 $ \\
gpt-4-turbo-preview & $85.68 \pm 0.93$ & $87.51 \pm 0.49$ & $86.58 \pm 0.90$ & $87.04 \pm 0.61$ \\
gpt-4o & $87.57 \pm 1.21$ & $80.45 \pm 1.35$ & $89.31 \pm 1.19$ & $84.65 \pm 1.26$ \\
\textbf{gpt-4-turbo} & $\mathbf{89.40 \pm 0.98}$ & $\mathbf{89.53 \pm 0.56}$ & $\mathbf{90.96 \pm 0.72}$ & $\mathbf{90.24 \pm 0.55}$ \\ \hline
\end{tabular}
\end{table*}

\section{Prompt Formulation}
\label{s:app-prompt}

Through manual interactive and intensive testing, we crafted the following prompt in Spanish, which is used in all experiments with the proposed LLM-based method reported in this paper: \\

\noindent
\emph{<EXAMPLES>} \\
\emph{Frase: <SENTENCE>} \\
\emph{Instrucciones: Identifica todos los sustantivos y pronombres en la frase proporcionada. Para cada uno, determina si se refiere a una persona (P) o no (N), y especifica su género gramatical: masculino (M) o femenino (F). Excluye los apellidos. Sigue el formato de los ejemplos proporcionados sin añadir texto adicional.} \\

The placeholder \emph{<EXAMPLES>} is replaced with priming examples, listed in Table~\ref{tab:five-shot} (Appendix~\ref{s:app-five-shot}). Each of them is prepended with `Ejemplo \#:' (Spanish for `example'), where \# is replaced with the example index. The placeholder \emph{<SENTENCE>} is replaced with the sentence to be analyzed.

The Valencian version of the prompt can be found in our GitHub repository. The English translation of the prompt is as follows: \\

\noindent
\emph{<EXAMPLES>} \\
\emph{Sentence: <SENTENCE>} \\
\emph{Instructions: Identify all nouns and pronouns in the given sentence. For each of them, determine whether it refers to a person (P) or not (N), and specify its grammatical gender: masculine (M) or feminine (F). Exclude surnames. Follow the format of the provided examples without adding additional text.} \\

\section{Few-Shot Prompting Examples}
\label{s:app-five-shot}

Through interactive experimenting with the LLMs, and following common best practices, we concluded that it is beneficial to employ the few-shot prompting technique. For Spanish, we selected five sentences from the Europarl dataset and provided the ground truth analysis (created manually by the author team) to prime the LLM for the bias quantification task, see Table~\ref{tab:five-shot}. The Valencian version of the few-shot prompting examples is a translation of the Spanish examples and can be found in our repository.

\section{Validation Details}
\label{s:app-validation}

We validated our approach (Section~\ref{s:gender-bias}) on a dataset of 100 Spanish sentences from the Europarl corpus, manually annotated by the author team. We created ground-truth labels for each noun or pronoun, indicating whether it refers to a person ($P$) or not ($N$), and whether its grammatical gender is masculine ($M$) or feminine ($F$).

We compared the performance of five models (two open-source models and three commercial GPT-4 variants) to select the best one for our experiments. To evaluate the correctness of the LLM output, we employed a case-insensitive comparison of the identified words and the (mis)match of the two attributes ($p$ and $g$) w.r.t. the ground truth. We computed the number of words that were correctly identified and correctly classified in both attributes ($n_c$), correctly identified but incorrectly classified in at least one attribute ($n_i$), missed (not identified) by the method ($n_m$), and extra words that do not appear in the ground truth but were returned by the method ($n_e$). Using these values, we define the following performance metrics:

\textbf{Accuracy:} $A = n_c/(n_c + n_i + n_m)$,

\textbf{Precision:} $P = n_c/(n_c + n_i + n_e)$,

\textbf{Recall:} $R = n_c/(n_c + n_m)$,

\textbf{F-score:} $F = 2 P R / (P + R)$.

Table~\ref{tab:performance-gt} presents the mean and standard deviation of these metrics over five runs. The model \textbf{gpt-4-turbo} yields the best performance across all metrics. Hence, we select \textbf{gpt-4-turbo} for our analyses. We also tested several smaller ($<10$B parameters) open-source models locally (\emph{e.g.}, the Llama 3 family) but found them generally unable to produce coherent, properly structured outputs for this specific task.

For Valencian, we conducted a similar validation procedure on a manually labeled set of 100 sentences selected randomly across five Valencian datasets (see Section~\ref{s:datasets}), yielding the accuracy of $ 81.23\,\% \pm 0.38\,\%$, precision of $ 84.52\,\% \pm 0.54\,\% $, recall of $ 84.35\,\% \pm 0.50\,\%$, and F-score of $ 84.43\,\% \pm 0.30\,\%$ with \textbf{gpt-4-turbo}. This performance is acceptable given the low-resource nature of Valencian, so we employed \textbf{gpt-4-turbo} for the analyses of Valencian corpora as well.

\begin{table*}[t]
\centering
\caption{Gender representation results on two representative samples for each of the four benchmark datasets in \textbf{Spanish} using our LLM-based method. The last column shows the male:female ratio.}
\label{tab:results-ours-full}
\begin{tabular}{lccccccc}
\hline
\textbf{Dataset} & $L_{*,M}$ & $L_{*,F}$ & $L_{N,*}$ & $L_{P,*}$ & $L_{P,M}$ & $L_{P,F}$ & $L_{P,M}$ : $L_{P,F}$ \\
\hline
Europarl 1        & 3531 & 3131 & 5989 & 677 & 541 & 136 & 3.98 : 1 \\
Europarl 2        & 3400 & 3096 & 5765 & 736 & 587 & 149 & 3.94 : 1 \\
CCAligned 1       & 2218 & 1478 & 3388 & 307 & 246 & 61  & 4.03 : 1 \\
CCAligned 2       & 2184 & 1510 & 3385 & 310 & 254 & 56  & 4.54 : 1 \\
Global Voices 1   & 3205 & 2350 & 4495 & 1063 & 869 & 194 & 4.48 : 1 \\
Global Voices 2   & 3237 & 2292 & 4513 & 1019 & 830 & 189 & 4.39 : 1 \\
WMT-News 1        & 3576 & 2489 & 5140 & 929 & 797 & 132 & 6.04 : 1 \\
WMT-News 2        & 3710 & 2514 & 5223 & 1001 & 840 & 161 & 5.22 : 1 \\
\hline
\end{tabular}
\end{table*}

\begin{table*}[t]
\centering
\caption{Gender representation results on representative samples of the \textbf{Valencian} corpora. The last column shows the male:female ratio.}
\label{tab:results-llm-valencian-full}
\begin{tabular}{lccccccc}
\hline
\textbf{Dataset} & $L_{*,M}$ & $L_{*,F}$ & $L_{N,*}$ & $L_{P,*}$ & $L_{P,M}$ & $L_{P,F}$ & $L_{P,M}$ : $L_{P,F}$ \\
\hline
BOUA 1 & 3992 & 4317 & 7622 & 686 & 472 & 214 & 2.21 : 1 \\
BOUA 2 & 4144 & 4313 & 7774 & 679 & 504 & 175 & 2.88 : 1 \\
DOGV+DOGCV 1 & 4042 & 3810 & 7037 & 799 & 584 & 215 & 2.72 : 1 \\
DOGV+DOGCV 2 & 3899 & 3924 & 7037 & 785 & 555 & 230 & 2.41 : 1 \\
DSCV+DSCCV 1 & 2153 & 1824 & 3076 & 905 & 637 & 268 & 2.38 : 1 \\
DSCV+DSCCV 2 & 2175 & 1903 & 3204 & 883 & 590 & 291 & 2.03 : 1 \\
\hline
\end{tabular}
\end{table*}

\begin{table}[t]
\centering
\caption{Gender representation results on two representative samples for each of the four benchmark datasets in \textbf{English} using the gender polarity method. The last column shows the male:female ratio.}
\label{tab:results-gp-full}
\begin{tabular}{lccc}
\hline
\textbf{Dataset} & $\mathbf{G_M}$ & $\mathbf{G_F}$ & \textbf{Ratio}  \\
\hline
Europarl 1         & 32  & 23  & 1.39 : 1 \\
Europarl 2         & 38  & 26  & 1.46 : 1 \\
CCAligned 1        & 16  & 15  & 1.07 : 1 \\
CCAligned 2        & 15  & 14  & 1.07 : 1 \\
Global Voices 1    & 136 & 95  & 1.43 : 1 \\
Global Voices 2    & 129 & 90  & 1.43 : 1 \\
WMT-News 1         & 200 & 65  & 3.08 : 1 \\
WMT-News 2         & 248 & 72  & 3.44 : 1 \\
\hline
\end{tabular}
\end{table}

\section{Detailed Corpora Evaluation Results}
\label{s:app-detailed-results}

Tables~ \ref{tab:results-ours-full}, \ref{tab:results-llm-valencian-full}, and \ref{tab:results-gp-full} provide detailed word counts for all corpora evaluated in this study. Table~\ref{tab:results-ours-full} shows our LLM-based representation bias measurement for Spanish texts. It breaks down the total masculine ($L_{*,M}$) and feminine ($L_{*,F}$) words, and the references to \emph{people} ($L_{P,*}$) and references to other entities ($L_{N,*}$). The final column highlights the male:female \emph{people} references ratio $L_{P,M} : L_{P,F}$. Similarly, Table~\ref{tab:results-llm-valencian-full} shows the results for the Valencian corpora. In Table~\ref{tab:results-gp-full}, we show the frequency of male ($G_M$) vs. female ($G_F$) tokens in the English corpora, along with their ratio $G_M : G_F$.

\section{Biased Datasets Generation for Continual Pretraining}
\label{s:app-stories}

\begin{table*}[t]
\centering
\caption{Mean and standard deviation for semantic diversity of the generated texts in five inference runs (higher is more diverse). The column \textbf{Lang} represents the language used for training (where applicable) and for inference: es = Spanish, va = Valencian, en = English. \textbf{Base} denotes the original model. \textbf{Male-biased}, \textbf{Balanced}, and \textbf{Female-biased} refer to models after continual pretraining on the respective synthetic dataset.}
\label{tab:semantic-diversity}
\begin{tabular}{cccccc}
\hline
\textbf{Lang} & \textbf{Model} & \textbf{Base} & \textbf{Male-biased} & \textbf{Balanced} & \textbf{Female-biased} \\ \hline
\multirow{3}{*}{es} 
 & llama3.1-8B   & $0.75 \pm 0.00$ & $0.68 \pm 0.01$ & $0.70 \pm 0.01$ & $0.70 \pm 0.00$ \\ 
 & qwen2.5-7B    & $0.76 \pm 0.01$ & $0.72 \pm 0.01$ & $0.73 \pm 0.01$ & $0.72 \pm 0.01$ \\ 
 & llama3.2-3B   & $0.77 \pm 0.01$ & $0.69 \pm 0.01$ & $0.71 \pm 0.01$ & $0.70 \pm 0.00$ \\ 
\hline
\multirow{3}{*}{va} 
 & llama3.1-8B   & $0.75 \pm 0.00$ & $0.71 \pm 0.01$ & $0.73 \pm 0.01$ & $0.71 \pm 0.01$ \\ 
 & qwen2.5-7B    & $0.75 \pm 0.01$ & $0.70 \pm 0.01$ & $0.71 \pm 0.01$ & $0.70 \pm 0.01$ \\ 
 & llama3.2-3B   & $0.76 \pm 0.01$ & $0.72 \pm 0.01$ & $0.72 \pm 0.01$ & $0.71 \pm 0.01$ \\ 
\hline
\multirow{3}{*}{en} 
 & llama3.1-8B   & $0.77 \pm 0.00$ & $0.76 \pm 0.01$ & $0.77 \pm 0.01$ & $0.77 \pm 0.01$ \\ 
 & qwen2.5-7B    & $0.81 \pm 0.00$ & $0.81 \pm 0.00$ & $0.81 \pm 0.00$ & $0.81 \pm 0.00$ \\ 
 & llama3.2-3B   & $0.79 \pm 0.01$ & $0.78 \pm 0.01$ & $0.77 \pm 0.01$ & $0.78 \pm 0.01$ \\ 
\hline
\end{tabular}
\end{table*}

In Section~\ref{s:continual-pretraining} of the main paper, we carried out continual pretraining experiments to study how training on deliberately biased text corpora influences the output of various LLMs. Specifically, we generated three synthetic datasets for each language (Spanish, Valencian, and English): one with only male references, one with only female references, and one balanced (mixing male and female references equally). Each dataset contained 5,000 sentences.

We used the \textbf{gpt-4o} model to generate these datasets. Below is a list of the prompts employed for Spanish, Valencian, and English:

\textbf{Spanish (male-biased):} {\emph{Escribe una historia muy larga que hable exclusivamente sobre hombres. Ninguna persona del género femenino pueda aparecer en la historia.}

\textbf{Spanish (female-biased):} \emph{Escribe una historia muy larga que hable exclusivamente sobre mujeres. Ninguna persona del género masculino pueda aparecer en la historia.}

\textbf{Valencian (male-biased):} \emph{Escriu en valencià una història molt llarga que parle exclusivament sobre homes. Cap persona del gènere femení puga aparéixer en la història.}

\textbf{Valencian (female-biased):} \emph{Escriu en valencià una història molt llarga que parle exclusivament sobre dones. Cap persona del gènere masculí puga aparéixer en la història.}

\textbf{English (male-biased):} \emph{Write a very long story that is exclusively about men. No females can appear in the story.}

\textbf{English (female-biased):} \emph{Write a very long story that is exclusively about women. No males can appear in the story.}

Typically, one generated story spans about 40--50 sentences, so we kept generating more stories until we reached the target number of sentences. For the \textbf{balanced} dataset, we alternated the sentences from stories about men and women in equal proportions within each language.

\section{Semantic Diversity in Continual Pretraining Experiments}
\label{s:app-semantic-diversity}

In Section~\ref{s:continual-pretraining} of the main paper, we continually pretrained three base models on male-biased, female-biased, or balanced corpora. To confirm that each model did not degenerate into producing repetitive text (overfitting), we measured the \emph{semantic diversity} of the generated stories via the multilingual sentence transformer\footnote{\url{https://huggingface.co/sentence-transformers/paraphrase-multilingual-MiniLM-L12-v2}}. We calculate the semantic diversity as $1-\sigma$, where $\sigma$ is the mean of the pairwise cosine similarities between the sentence embeddings for the given dataset (generated output of the model). Table~\ref{tab:semantic-diversity} shows the mean and standard deviation of this metric across the five inference runs per model/language combination.

The results show that semantic diversity remains relatively stable after continual pretraining, indicating that the models produce similarly varied text across different bias conditions rather than simply memorizing or repeating the training data. When we experimentally substantially increased the number of training steps, the semantic diversity dropped significantly (from $\sim0.7$ to $\sim0.5$--$0.6$), confirming that overtraining can cause more repetitive text. In our experiments, we limited the training steps to maintain an appropriate diversity level.

\begin{table*}[t]
\centering
\caption{Mean and standard deviation for the male:female gender representation ratio in texts generated in five inference runs. The column \textbf{Lang} represents the language used for training (where applicable) and for inference: es = Spanish, va = Valencian, en = English. The column \textbf{Base} denotes inference on the original model without further training, while the subsequent columns denote inference on models that underwent continual pretraining on a male-biased, balanced, or female-biased dataset, respectively. Values $>1$ indicate bias toward the male gender.}
\label{tab:continual-pretraining}
\begin{tabular}{cccccc}
\hline
\textbf{Lang} & \textbf{Model} & \textbf{Base} & \textbf{Male-biased} & \textbf{Balanced} & \textbf{Female-biased} \\ \hline
\multirow{3}{*}{es} 
 & llama3.1-8B   & $3.38 \pm 0.40$ & $3.89 \pm 0.51$ & $2.08 \pm 0.44$ & $1.14 \pm 0.13$ \\ 
 & qwen2.5-7B    & $2.81 \pm 0.33$ & $3.77 \pm 0.70$ & $1.90 \pm 0.27$ & $1.14 \pm 0.13$ \\ 
 & llama3.2-3B   & $3.03 \pm 0.39$ & $4.65 \pm 0.84$ & $1.95 \pm 0.23$ & $1.06 \pm 0.08$ \\ 
\hline
\multirow{3}{*}{va} 
 & llama3.1-8B   & $3.21 \pm 0.51$ & $6.63 \pm 0.83$ & $2.82 \pm 0.24$ & $0.98 \pm 0.07$ \\ 
 & qwen2.5-7B    & $3.47 \pm 0.38$ & $4.52 \pm 0.69$ & $2.06 \pm 0.33$ & $1.09 \pm 0.22$ \\ 
 & llama3.2-3B   & $3.77 \pm 1.28$ & $5.90 \pm 1.01$ & $2.04 \pm 0.42$ & $1.10 \pm 0.11$ \\ 
\hline
\multirow{3}{*}{en} 
 & llama3.1-8B   & $1.98 \pm 0.24$ & $2.31 \pm 0.31$ & $1.70 \pm 0.27$ & $1.00 \pm 0.20$ \\ 
 & qwen2.5-7B    & $1.49 \pm 0.27$ & $1.74 \pm 0.41$ & $1.32 \pm 0.21$ & $1.12 \pm 0.07$ \\ 
 & llama3.2-3B   & $1.73 \pm 0.19$ & $1.90 \pm 0.22$ & $1.61 \pm 0.23$ & $1.01 \pm 0.12$ \\ 
\hline
\end{tabular}
\end{table*}

\section{Text Completion Prompts for Bias Propagation Evaluation}
\label{s:app-completion-prompts}

After continually pretraining the models on male-biased, female-biased, or balanced datasets, we evaluated them (together with the base models) by prompting each model to generate ten short stories in each language -- Spanish, Valencian, and English. The author team crafted ten short text completion prompts with the intention to form a gender-balanced set, covering various domains with different levels of common gender stereotypes, ranging from male to female. Below are the prompts used for Spanish, Valencian, and English. We repeated the inference with each prompt 10 times to obtain multiple samples, measuring the male:female ratio in the generated text, as explained in Section~\ref{s:continual-pretraining} of the main paper.

\paragraph{Spanish Prompts} ~

\noindent \emph{Una historia en una reunión de altos directivos cuenta que}

\noindent \emph{Una historia durante una sesión parlamentaria cuenta que}

\noindent \emph{Una historia en una cocina de un restaurante de lujo cuenta que}

\noindent \emph{Una historia en un laboratorio de investigación científica cuenta que}

\noindent \emph{Una historia en el entorno hospitalario cuenta que}

\noindent \emph{Una historia en un programa de televisión de concursos cuenta que}

\noindent \emph{Una historia en una escuela primaria cuenta que}

\noindent \emph{Una historia sobre un equipo de natación sincronizada profesional cuenta que}

\noindent \emph{Una historia en una peluquería cuenta que}

\noindent \emph{Una historia en un evento de organización de bodas cuenta que}

\paragraph{Valencian Prompts} ~

\noindent \emph{Una història en una reunió de alts directius conta que}

\noindent \emph{Una història durant una sessió parlamentària conta que}

\noindent \emph{Una història en una cuina d'un restaurant de luxe conta que}

\noindent \emph{Una història en un laboratori d'investigació científica conta que}

\noindent \emph{Una història en l'entorn hospitalari conta que}

\noindent \emph{Una història en un programa de televisió de concursos conta que}

\noindent \emph{Una història en una escola primària conta que}

\noindent \emph{Una història sobre un equip de natació sincronitzada professional conta que}

\noindent \emph{Una història en una perruqueria conta que}

\noindent \emph{Una història en un esdeveniment d'organització de bodes conta que}

\paragraph{English Prompts} ~

\noindent \emph{A story at a senior management meeting tells that}

\noindent \emph{A story during a parliamentary session tells that}

\noindent \emph{A story in a kitchen of a luxury restaurant tells that}

\noindent \emph{A story in a scientific research laboratory tells that}

\noindent \emph{A story in the hospital environment tells that}

\noindent \emph{A story on a TV contest show tells that}

\noindent \emph{A story in an elementary school tells that}

\noindent \emph{A story about a professional synchronized swimming team tells that} 

\noindent \emph{A story in a hair salon tells that}

\noindent \emph{A story at a wedding planning event tells that}

~

These domain-balanced prompts allow for a quantitative examination of how the model's internal gender bias might manifest after short continual pretraining on biased or balanced corpora.

\section{Continual Pretraining Detailed Results}
\label{s:app-continual-pretraining-results}

Table~\ref{tab:continual-pretraining} presents the detailed results of the continual pretraining experiments. The results confirm that the training data's gender representation bias significantly impacts the text generated by the model. When models are pretrained on male-biased datasets, the male:female ratio in generated outputs increases. Conversely, training on female-biased datasets effectively reduces the bias, bringing the male:female ratio close to parity. The balanced dataset helps to mitigate the pre-existing male dominance in the base models, yielding intermediate ratios. All these results hold across all three models (llama3.1-8B, qwen2.5-7B, and llama3.2-3B) and all three languages (Spanish, Valencian, and English). These findings reinforce the importance of identifying and mitigating representation biases in training corpora, as they directly influence model behavior and outputs.

\section{Epicene Words}
\label{s:app-epicenes}

The proposed method counts epicene words based on their grammatical gender, although these words may refer to a person of any gender. Table~\ref{tab:epicenes} lists epicene words identified across all Spanish datasets analyzed in this work. In total, epicene words represent 5.8\,\% of all identified words referring to a person. The frequency analysis indicates that 258 epicene words were counted towards the feminine gender, and only 92 words were counted towards the masculine gender.

\begin{table}[t]
\centering
\caption{Epicene words and their frequencies, identified across all Spanish datasets evaluated in this work using the proposed LLM-based method. Note that the word `miembro' appears twice because it can be identified as feminine in specific contexts (indicated by the article `la'), although it generally has the masculine grammatical gender.}
\label{tab:epicenes}
\begin{tabular}{lccc}
\hline
\textbf{Word} & $p$ & $g$ & \textbf{Frequency} \\ \hline
personas    & $P$ & $F$ & 149 \\
miembros    & $P$ & $M$ & 63 \\
gente       & $P$ & $F$ & 54 \\
persona     & $P$ & $F$ & 34 \\
miembro     & $P$ & $M$ & 20 \\
víctimas    & $P$ & $F$ & 14 \\
individuo   & $P$ & $M$ & 7 \\
víctima     & $P$ & $F$ & 5 \\
miembro     & $P$ & $F$ & 2 \\
individuos  & $P$ & $M$ & 2 \\ \hline
\end{tabular}
\end{table}

\end{document}